\documentclass{article}
\pdfoutput=1

\usepackage{arxiv}

\usepackage[utf8]{inputenc} 
\usepackage[T1]{fontenc}    
\usepackage{lmodern}	    
\usepackage{hyperref}	    
\usepackage{titlesec}       
\usepackage{url}	    
\usepackage{booktabs}	    
\usepackage{amsfonts}	    
\usepackage{nicefrac}	    
\usepackage{microtype}	    
\usepackage{cleveref}	    
\usepackage{moreverb}       
\usepackage{lipsum}	    
\usepackage{enumitem}
\usepackage{listings}
\usepackage{graphicx}
\graphicspath{ {./pics/} }
\usepackage{natbib}
\setcitestyle{square}
\setcitestyle{numbers}
\usepackage{doi}
\usepackage[table,xcdraw]{xcolor}
\usepackage{footnote}
\makesavenoteenv{tabular}
\makesavenoteenv{table*}

\providecommand{\keywords}[1]
{
	\small
	\textbf{\textit{Keywords---}} #1
}

\lstnewenvironment{verbquote}[1][]
  {\lstset{columns=fullflexible,
           basicstyle=\ttfamily,
           xleftmargin=2em,
           xrightmargin=2em,
           breaklines,
           breakindent=0pt,
           #1}}
  {}
 
\title{NSOAMT - New Search Only Approach to Machine Translation}

\date{September 19, 2023}

\usepackage{authblk}

\author[1]{
Jo\~{a}o~Lu\'{\i}s
}
\author[1]{
	Diogo~Cardoso
}
\author[1]{
Jos\'{e}~Marques
}
\author[1]{
Lu\'{\i}s~Campos
}
\affil[1]{nsoamt.team at \href{https://www.pdmfc.com}{pdmfc.com}}

\hypersetup{
	pdftitle={NSOAMT - New Search Only Approach to Machine Translation},
	pdfsubject={NLP},
	pdfauthor={Luís Campos, João Luís, Diogo Cardoso, José Marques},
	pdfkeywords={NSOAMT, NLP,Natural language processing, Translation, Text metrics}
}

\begin{document}
\maketitle

\footnotetext[1]{This work was co-sponsored by Lisboa2020 / Portugal2020 project LISBOA-01-0247-FEDER-045071.}

\begin{abstract}
    Translation automation mechanisms and tools have been developed for several years to bring people who speak different languages together. 
    A "new search only approach to machine translation" was adopted to tackle some of the slowness and inaccuracy of the other technologies. The idea is to develop a solution that, by indexing an incremental set of words that combine a certain semantic meaning, makes it possible to create a process of correspondence between their native language record and the language of translation.
    This research principle assumes that the vocabulary used in a given type of publication/document is relatively limited in terms of language style and word diversity, which enhances the greater effect of instantaneously and rigor in the translation process through the indexing process.
    A volume of electronic text documents where processed and loaded into a database, and analyzed and measured in order confirm the previous premise.
    Although the observed and projected metric values did not give encouraging results, it was possible to develop and make available a translation tool using this approach.

\end{abstract}


\section{Introduction}
\label{sec:intro}

Translation automation mechanisms and tools have been developed for several years to bring people who speak different languages together. In the last year most of these tools have been based on deep learning, in part due to the rise of AI technologies, but also due to some abstraction it provides to the multiple language semantics that exists.
 
In this paper we describe a research project, named {\em New Search Only Approach to Machine Translation} (NSOAMT) developed to tackle some of the issues (inaccuracies, etc.) of the other approaches. The idea is to develop a solution that, by indexing an incremental set of words that combine a certain semantic meaning, makes it possible to create a process of correspondence between their native language record and the language of translation.

This research principle assumes that the vocabulary used in each type of publication/document is relatively limited in terms of language style and word diversity, which enhances the greater effect of instantaneously and rigor in the translation process through the indexing process.

In this paper we present the results we found when putting such principles to practice, as we attempt build a machine translation service based on such premises.

\section{Problem statement}
\label{sec:problem}

Although several general purpose language translation services already exist, it is still known that for high-quality translations in specific domains, a human expert is still required (\cite{wang2021progress}, \cite{wiesmann2019machine}, \cite{jinfang2023exploring}).

Natural language sentences are just a sequence of words. In the eventuality that we could quantify and store the most commonly used sentences, could this domain expertise for machine translation purposes be crowd-sourced (\cite{enwiki:1172855974})?

If not, can the distance to this practical goal be measured (or at least guessed) ?

\section{State of the art}
\label{sec:stateoftheart}

The evolution of natural language processing has come a long way since the early 1950's.\cite{johri2021natural}. There are many technological approaches, but these can be mostly split into three major categories \cite{enwiki:1173873917}:
\begin{enumerate}
    \item symbolic rule-based processing systems
    \item statistical approaches
    \item neural network based 
\end{enumerate}

Although the present's focus is mostly on neural network based techniques (due to the popularity of Large Language Models \cite{enwiki:1173840397}), the approach discussed in this article is best categorized as a "statistical approach".

\section{Methodology}
\label{sec:methodology}

\begin{enumerate}
    \item Import vast quantities of text documents, broken into sentences.
    \item Identify common text fragments.
    \item Crowdsource \cite{enwiki:1172855974} translation of common text fragments.
\end{enumerate}

This methodology is only viable today due to:
\begin{itemize}
    \item The Internet and the World Wide Web connecting several organizations, institutions, and private initiatives, making available several sources of text. (See sec.\ref{sec:textsources}).
    \item General availability of open-source libraries and tools, such as NLTK, that enable quick prototyping of some NLP techniques. (See sec.\ref{sec:nltk}).
    \item The consequences of the Moore's law \cite{enwiki:1172581333} allowing for hardware capable of handling Terabytes of text at a reasonable cost.
\end{itemize}

\subsection{Sentence model}
\label{sec:sentencemodel}

\begin{figure*}[htbp]
	\centering
	\includegraphics[width=0.7\textwidth]{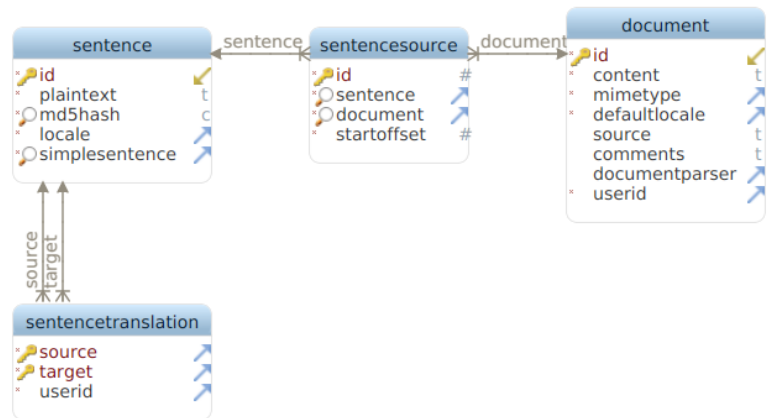}
	\caption{Entity-relationship sentence diagram.}
	\label{fig:ersentence}
\end{figure*}

Figure~\ref{fig:ersentence} illustrates the relational model for the documents and sentences ingested. The description of each table is:

\begin{description}
    \item[document] - one table row for each ingested document. The {\tt content} column contains the UTF-8 plain text of the document before sentence separation.
    \item[sentence] - one table row for each distinct sentence. The {\tt plainText} column contains the UTF-8 text of the sentence, including punctuation and inner spaces. There are no duplicate sentences for the same {\tt plainText} content, so it can be said that the rows in table sentence represent {\bf distinct sentences}. (A count of rows on table sentence in a given context is the value of the metric of {\bf \#distinct sentences} in that same context).
    \item[sentencesource] - one table row for sentence extracted from the document. The column {\tt startOffset} is the sentence sequence order number. (A count of rows on table sentencesource in a given scope is the value of the metric of {\bf \#sentences} in that same scope).
    \item[sentencetranslation] - one table row for a possible (context-free) translation of a sentence.
\end{description}

\subsection{Metrics}
\label{sec:metrics}

When trying to translate a document it is assumed that there is already a
database that contains all possible translations for each of the
the sentences in the document.
Logically, it can be assumed that the more sentences
that are imported, the better chances there are of obtaining a correspondence
between a new sentence and a sentence that already exists in the system.
This line of thought thus allows measuring the current state of
the system with the import of more and more files using the following metrics:

\begin{description}
	\item[\#sentences] - count of sentences in the text, as parsed by a sentence tokenizer software. The sentence separation is not as accurate as a human reader would perceive it, but an effort was made to make it consistent across all measurements. See sec.\ref{sec:nltk}.
	\item[\#distinct sentences] - (sometimes abbreviated as {\bf \#d.sentences}) How many distinct sentences exist in
		system and information of the idea of the distance at which
		the theoretical
		ceiling is found; These can be divided into subcategories:
		\begin{description}
			\item[\#distinct sentences without repetitions] - (also called {\bf \#unique d.sentences} for shortness) How many distinct
				sentences exist that have been referenced (used
				in a document) only once in the
				whole sentence database. (In the short "\#unique d.sentences", the "d." standing for "distinct" is redundant, as all unique sentences are distinct, but makes clearer that it should be related with the "\#distinct sentences" metric, and not the "\#sentences" metric).
			\item[\#distinct sentences with repetitions] - How many distinct sentences have
				been referenced more than once (regardless of the document, and regardless of
				the multiplicity of the repetition).
		\end{description}
\end{description}

For illustration purpose, Figure~\ref{fig:exampletext} has a small example text document, which results in the metrics shown in Table~\ref{tab:examplemetrics}.

\begin{figure*}[htbp]
	\centering
	\begin{boxedverbatim}
When parrots do it, it's parroting.
When children do it, it's imitation.
When computers do it, it's AI.
When parrots do it, it's parroting.
\end{boxedverbatim}
	\caption{Example text document.}
	\label{fig:exampletext}
\end{figure*}

\begin{table*}[htbp]
	\centering
\begin{tabular}{l||r}
\hline
                                     & Example (en) \\ \hline \hline
\#documents                          & 1            \\ \hline
\#text characters (UTF-8)            & 140          \\ \hline
\#sentences                          & 4            \\ \hline
\#distinct sentences                 & 3            \\ \hline
\#distinct sentences \%              & 75\%         \\ \hline
\#d.sentences with repetitions       & 1            \\ \hline
\#d.sentences with repetitions \%    & 33,33\%      \\ \hline
\#unique d.sentences                 & 2            \\ \hline
\#unique d.sentences \%              & 66,67\%      \\ \hline
\#non-unique sentences \%              & 50,00\%      \\ \hline
\end{tabular}
	\caption{Metrics for the example text in Figure \ref{fig:exampletext}}
	\label{tab:examplemetrics}
\end{table*}

Description of the other metrics shown in Table~\ref{tab:examplemetrics}:

\begin{description}
    \item[\#text characters] - include line breaks and other invisible characters, so it might vary for the same visual text content. For the English language texts, it is also a good approximation of the metric of the volume of information bytes processed, as UTF-8 character encoding \cite{enwiki:1171220477} is used.
    \item[\#distinct sentences \%] - the percentage is calculated as $\frac{\#distinct\  sentences}{\#sentences}$
    \item[\#unique d.sentences \%] - the percentage is calculated as $\frac{\#unique\  d.sentences}{\#distinct\ sentences}$.
    \item[\#non-unique sentences \%] - You can calculate the percentage of sentences with repetitions using the expression $\frac{\#sentences - \#unique\ d.sentences}{\#sentences}$. For this example, it is 50\%. This metric is not usually shown (as the ratio between the two underlying values can be easily observed), but should not be compared with the {\bf \#unique d.sentences \%} metric.
\end{description}

\subsection{Theoretical limits}
\label{sec:upperbound}

The feasibility of the project
assumes that the vocabulary in use is limited. Based on this,
it can be assumed that the number of feasible sentences resulting
from possible vocabulary combinations is also limited. However, this claim
contradicts the opinion of linguists who often point to the potential for an
infinite number of possible sentences to exist.

It is also known that for an infinite number of sentences to exist, at least one of the following conditions must be met:

\begin{itemize}
    \item There are an infinite number of words;
    \item A sentence can contain any number of words (unbounded).
\end{itemize}

For the first condition (there are an infinite number of words) to hold, there would have to be an infinite number of symbols (letters) or the words have an infinite number of letters. In western languages the number of symbols is not infinite and the longest word in the world contains 189819 letters \cite{devlin2021longest} (being, however, a technical word not used in natural language) so it can be admitted that there is a finite number of words, because none of the conditions is verified.
It is true that new words that exceed this limit can be created, but it is also possible to admit that the number of words that will be created is also finite and that these same words will not be used in everyday life. In this way, it is feasible to admit that there is a finite number of words used within an infinite number of possible words \cite{rgdisc}.

To estimate the number of existing words, the Oxford dictionary, which contains about 600,000 words, can be used as a basis. This number is constantly increasing; however the authors also assume that the number of new words will grow in a finite way, in the same way that the number of archaic words (i.e. words that are not used) also grows.

The second condition (a sentence contains any number of words) also holds, as shown by the examples found \cite{levin2020longest}. Obviously, these examples are exceptions and in common communication longer sentences have a lower level of understand-ability. To answer the question of “How many words can a sentence contain in order to maintain effective communication?” It is possible to find studies that point out that, from 43 words, a reader only understands 10\% of the content. For this reason, some organizations (such as the UK Government) recommend a maximum limit of 25 words per sentence to maintain effective communication (\cite{vincent2014length}).

Based on the previous numbers, a maximum limit was estimated for the universe of sentences with comprehensibility above 10\%, increasing the number of words in the dictionary to the number of words in a sentence. Thus, the number of possible sentences was limited to:

\begin{equation}
\sum_{n=1}^{n=43}{600000}^n\approx{600000}^{43}\approx288.74\times{10}^{246}
\end{equation}

This value is a theoretical ceiling, as it is not possible to randomly combine 43 words and generate, in all iterations, a grammatically correct sentence. Estimating the possible number of grammatically correct sentences is extremely complex, because, to do so, one would have to understand it in such a way that so that it would be possible to enumerate them.

According to a 1953 work by Michael West, it was concluded that, out of 600,000 words, it is possible to create a list with approximately 2000 words that represent a coverage of 80\% of the entire text, often written. This list was published under the name of “General Service List” (G.S.L.) \cite{enwiki:1170308086}. In 2013 (60 years after the creation of the original list) the list was expanded to contain 2818 words and is published under the name of “New General Service List” (N.G.S.L.) \cite{enwiki:1156325155}. This new list increased the coverage of the entire text to around 90\%. Given this new information, it was possible to repeat the calculation, with a view to trying to cover the maximum amount of text, with the fewest possible sentences:

\begin{equation}
\sum_{n=1}^{n=43}{2818}^n\approx{2818}^{43}\approx22,26\times{10}^{147}
\end{equation}

Again, this represented a theoretical ceiling, being practically lower, for the same reason described above. Limiting this value to the advised 25 words, the universe of possible phrases is even smaller:

\begin{equation}
\sum_{n=1}^{n=25}{2818}^n\approx{2818}^{25}\approx177,22\times{10}^{84}
\end{equation}

These 2818 words only represent the text written in everyday life. As the vocabulary used is circumstantial, when entering a context, new words will have to be added to obtain the same level of coverage. With this motivation, 3 new lists were created, which do not repeat words with the N.G.S.L.:

\begin{itemize}
    \item 960 words - "Academic Word List" (N.A.W.L.)\cite{nawl}: 92\% coverage;
    \item 1200 words - "TOEIC Service List" (T.S.L.)\cite{tsl}: 99\% coverage;
    \item 1700 words - "Business Word List" (B.S.L.)\cite{bsl}: 97\% coverage.
\end{itemize}

\begin{table*}[htbp]
\centering
\begin{tabular}{l||r|r|r}
\hline
List of Words & \# total of words & Ceiling for 25 words & Ceiling for 43 words \\ \hline
N.A.W.L. & 3778 & $2.70\times{10}^{89}$ & $6.64\times{10}^{153}$ \\
T.S.L. & 4018 & $1.25\times{10}^{90}$ & $9.38\times{10}^{154}$ \\
B.S.L. & 4518 & $2.36\times{10}^{91} $ &  $2.36\times{10}^{91}$  \\ \hline
\end{tabular}
	\caption{Descriptive table of the number of words per word list and maximum possible combinations for advisable sentence length (25 words) and sentence length where it is incomprehensible (43 words)}
	\label{tab:wordlist}
\end{table*}

Therefore, Table~\ref{tab:wordlist} presents the limits of possible sentences, using the previous lists, as a "theoretical limit". (The number of possible sentences that "make sense" is expected to be lower).

Given the above, it became necessary to verify the assumption, starting by importing enough sentences, which would allow obtaining a satisfactory degree of correspondence, as well as having a projection of necessary sentences, below the theoretical maximum limit.

\section{Implementation}
\label{sec:implementation}

This section describes the software stack (both technology, and implementation design choices), used to carry out the measurements and implement the resulting web site.

\subsection{Text sources}
\label{sec:textsources}

The text sources used where:

\begin{description}
    \item[\url{https://eur-lex.europa.eu/}] - Legislation documents from the European Union, available in 24 languages; HTML format.
    \item[\url{https://dumps.wikimedia.org/}] - Wikipedia backup dumps. XML+Wikitext format.
    \item[\url{https://arxiv.org/}] - Open-access scholarly articles. PDF format.\footnote{Non-PDF articles where discarded.} Download was performed my mirroring tools, with articles organized in monthly folders. (Only the latest version of each article was ingested.)
    \item[tBooks] - Several plain text literature content, obtained from sources like \url{https://www.gutenberg.org/}, \url{https://chroniclingamerica.loc.gov/},  
    \url{https://muse.jhu.edu/},
    \url{https://market.cantook.com/},
    \url{https://www.bookrix.com/},
    \url{https://archive.org/},
    \url{https://manybooks.net/},
    \url{https://www.smashwords.com/},
    \url{http://digital.library.upenn.edu/books/}. Plain text (UTF-8) format. We call this source aggregate {\bf tBooks}.\footnote{The content extracted from these sources is not publicly accessible on the NSOAMT site, due to possible copyright issues.}
\end{description}

\subsection{Ingestion pipeline}
\label{sec:ingestion}

\begin{figure*}[htbp]
	\centering
	\includegraphics[width=0.3\textwidth]{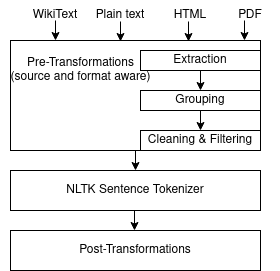}
	\caption{Ingestion pipeline general structure.}
	\label{fig:ipipeline}
\end{figure*}

The first stage of the ingestion pipeline is loading a content (in a specific electronic format) and splitting into plain text sentences (see Figure~\ref{fig:ipipeline}).

The actual sequence and details of each transformation depend on the format and source of the text. See \ref{sec:eformats} for issues and caveats of each source/format.

\begin{figure*}[htbp]
	\centering
	\includegraphics[width=0.7\textwidth]{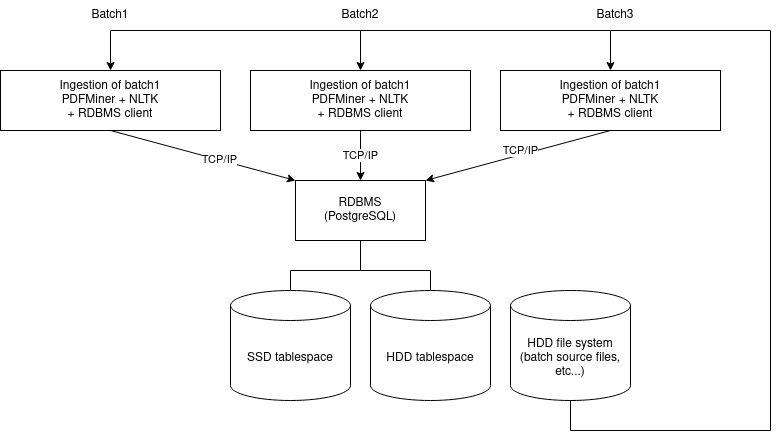}
	\caption{Concurrent PDF batch ingestion.}
	\label{fig:pdfbatch}
\end{figure*}

The last stage of the ingestion pipeline is loading the batches of parsed documents into the database. For a large source, such as arXiv, concurrent/parallel loading was needed, as shown in Figure~\ref{fig:pdfbatch}.

The high level algorithm for ingestion is:
\begin{enumerate}
    \item Format conversion to (HTML, WikiText, PDF, etc.) to plain text (or plain text groups).
    \item Split text into sentences (using \ref{sec:nltk}). Apply sentence transformation procedures (such as hash calculation). 
    \item insert the whole document into the database.
    \item for each sentence (by order of occurrence in the document):
    \begin{enumerate}[label*=\arabic*.]
        \item search if the sentence already exists in the database:
        \begin{enumerate}[label*=\arabic*.]
            \item if yes, associate existing sentence to current document.
            \item if no, insert the new sentence into the database, and then associate to the current document.
        \end{enumerate}
    \end{enumerate}
    \item (Post ingestion) Duplicate sentence elimination.
\end{enumerate}

Steps 1 and 2 can be sped up using multiple processes (when each document and sentences fit into memory). Steps 3 and 4 are performed in a single transaction (to avoid having non-parsed documents in the database), and can also be sped up using parallel execution, but there is a race condition between steps 4.1.1. and 4.1.2. Hence the need for a post-ingestion duplicate clean-up on step 5.\footnote{Normally, in a SQL database, the race condition would be avoided using a {\tt UNIQUE INDEX} on the column {\tt sentence.plainText}. But occasionally, long sentences are ingested, bumping into an undocumented PostgreSQL v12 limitation of unique indexes on large text values (which failed above 2704 plain text bytes, well below the expected 8129 bytes index page size limit).} 

\subsubsection{md5hash}
\label{sec:hash}

The data model exhibited in Figure~\ref{fig:ersentence} shows a column {\tt sentence.md5hash}, added for (read) indexing purposes. (As SQL built-in indexing was not possible\footnote{
Abnormally long sentences would not be indexed by regular PostgreSQL v12 B-Tree INDEX. And neither full-text-search, trigrams, or inverted indexes work with these uncommonly large text strings, for that matter.} due to very long sentences. These long "plain text" sentences are probably {\em garbage} resulting from a bad text extraction, but the decision was made to keep them, for future research).

The choice of the hashing algorithm to be MD5 \cite{enwiki:1171105697} (between MD5, SHA1, and others) were based on small storage requirements, speed of calculation, and the fact that implementations in python v3.6 hashlib and PostgreSQL v12 gave identical results.

It is known that the MD5 algorithm has a lower collision resistance (when compared to other more recent algorithms) \cite{enwiki:1171091464}, but as the purpose here was just to sped up the search (not cryptography grade collision resistance), it suffices. Note that in the model \ref{sec:sentencemodel} hash collisions are possible, expected, and well handled.

\subsubsection{Electronic document formats}
\label{sec:eformats}

\paragraph{Plain text}

UTF-8 encoded text documents.\cite{enwiki:1171220477}

Pros: Simpler to inspect and compare resulting model data to original text.

Cons: Separation between text fragments is sometimes not clear. Example: Titles, page headers, footers and foot notes, sentences interrupted by line and page breaks, sometimes placed and mixed amongst the text content without any convention consistency.
Sometimes it is possible to develop a special transformation that identifies these occurrences and splits the text into blocks (without these issues). Sometimes not. (Usually not done, because conventions vary a lot between documents, even from the same source).

\paragraph{WikiText}

Conversion to plain text done using~\cite{wikiextractor}.

Pros: Same as plain text. Additionally, the structure of Wikipedia's extracted text (and text style) splits into sentences very well, using NLTK's default sentence tokenizer API.

Cons: None that came to mind, although is a relatively short source (in terms of available volume of text).

\paragraph{Hyper Text Markup Language (HTML)}

Extraction of text from HTML 
\cite{enwiki:1171458148} content is done using \cite{beautifulsoup}.

Pros: Block tags force separation of text fragments (forcing sentence breaks).

Cons: Consistency of the formatting (and tag use) in the content layout is a very case-by-case approach. Page layout and navigation information filtering is also handled on a specific source-by-source case.

\paragraph{Portable Document Format (PDF)}

Text extraction from PDF files \cite{enwiki:1171241740} is performed using pdfminer.six.~\cite{pdfminer}

Pros: Largest source volume of documents available (example: arXiv).

Cons: Extraction of text from scientific articles in PDF format is problematic (\cite{yu2020extracting} and \cite{rama2012}). This results in many badly broken sentences. Some PDFs files have internal text represented in ways that result in garbled extracted text, while others even break the extraction process (and as such, PDF file upload is not publicly available on the NSOAMT site).

\subsubsection{Sentence tokenizer}
\label{sec:nltk}

NLTK \cite{nltk} is a python framework for natural language processing. The default sentence tokenizer API was used to extract an ordered list of sentences from plain text content. It is the core of the ingestion pipeline.

Using the default API (without a custom tokenizer trained for a specific text style) does not always produces good results (specifically in text extracted from scientific article's), but the same tokenizer results were consistent across all measurements.

\subsection{Sentence validation}
\label{sec:languagetool}

By sampling a few sentences, several examples with unusual (incorrect) grammar are easily spotted:

\begin{verbquote}
...
, 2017, 1-10 Editors: Will be set by the publisher 7 1 0 2 v o N 7 ] G L .

s c [ 3 v 1 0 0 0 0 . 

2 0 7 1 : v i X r a LEARNING THE DISTRIBUTION WITH LARGEST MEAN: TWO BANDIT FRAMEWORKS * Emilie Kaufmann 1 and Aur'elien Garivier 2 Abstract.
...
\end{verbquote}

As it stands now, the system includes a lot of partial sentences resulting from issues like text mis-extraction and sentence tokenization on titles, headers, footers, formulas, tabular data, graphics extracted as text, graphical text overlays, etc.

There are at least two possible mitigation strategies for this problem:

\begin{itemize}
    \item Improve the quality of the text extraction (and sentence tokenization).
    \item Exclude the {\em improperly} extracted sentences from the metrics.
\end{itemize}

Improving the quality of text extraction and sentence tokenization seems a never-ending battle (recognize/learn/train/develop text extractors for a never-ending variety of distinct specific documents and text styles). 
As such, the efforts were focused on filtering out {\em improperly} extracted sentences (because it simply felt a simpler and smaller task).

The "LanguageTool" version 5.7 \cite{ltool} seemed like a good candidate for an out-of-the-box linguistic tool that classifies sentences as valid or non-valid: It is open-source, suitably licensed, can be used on-premises and off-line, has a python interface language-tool-python, and the set of validation rules can be customized.

In the results section (\ref{sec:results}), sentences that have been checked using this tool are referred to as {\bf valid sentences}.

Note that, filtering sentences using such a tool, provides no warranty that we are working with a subset of "commonly used English" sentences that make sense (when read by a human). It just eliminates a large number of sentences that contain grammar rule violations (from the set of rules that the tool implements), some of which may be caused by bad text extraction, others by bad sentence tokenization, amongst other causes.

\subsection{Web interface}
\label{sec:webinterface}
\begin{figure*}[htbp]
	\centering
	\includegraphics[width=0.7\textwidth]{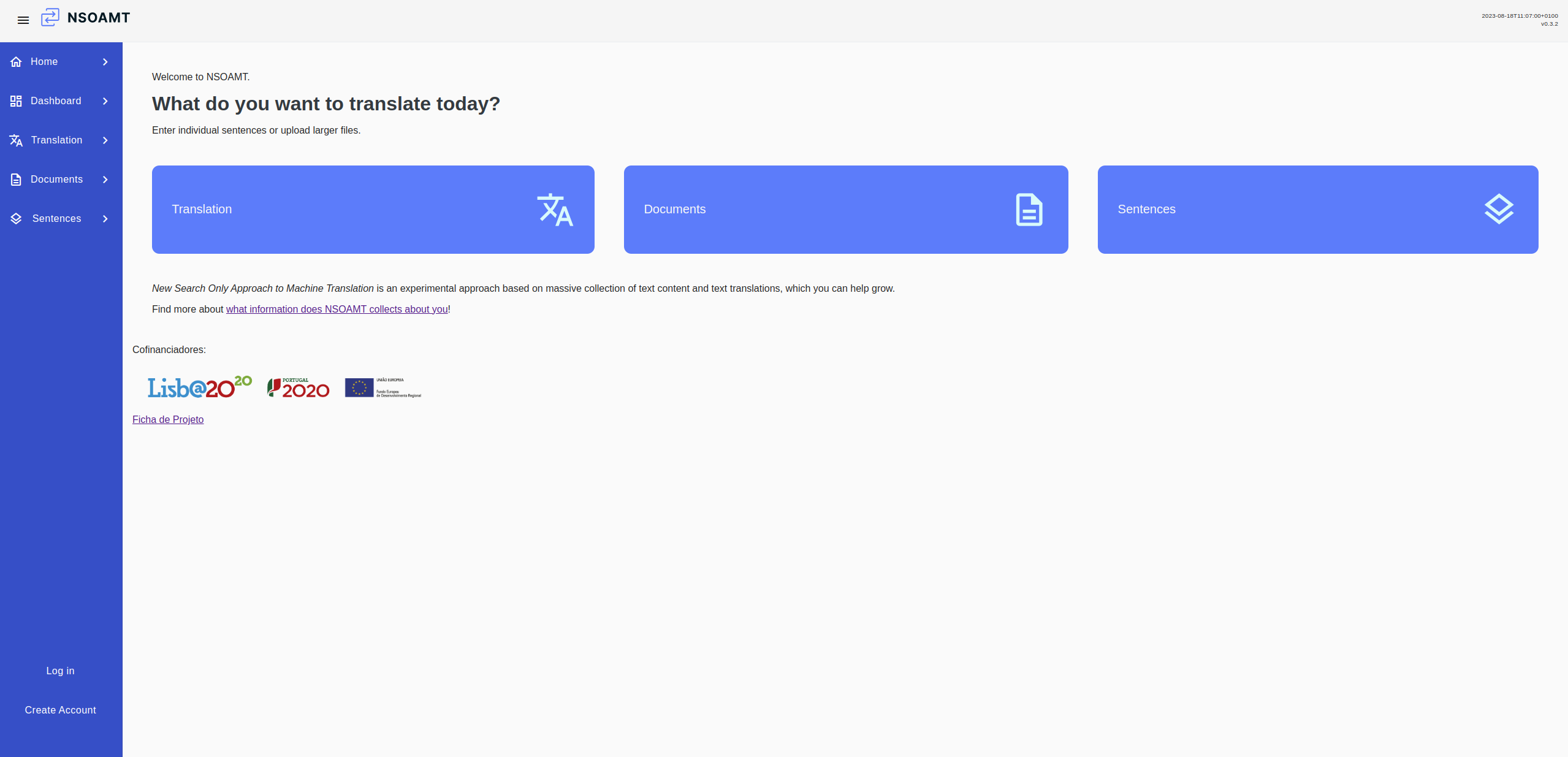}
	\caption{Application landing page.}
	\label{fig:landingpage}
\end{figure*}

In order to share the knowledge developed and show the planned translation functionality, a web interface for translation based on the technology studied was developed and made publicly available. This site is available at \href{https://nsoamt.pdmfc.com}{https://nsoamt.pdmfc.com}.
See Figure \ref{fig:landingpage} for the web frontend homepage.

\begin{figure*}[htbp]
	\centering
	\includegraphics[width=0.7\textwidth]{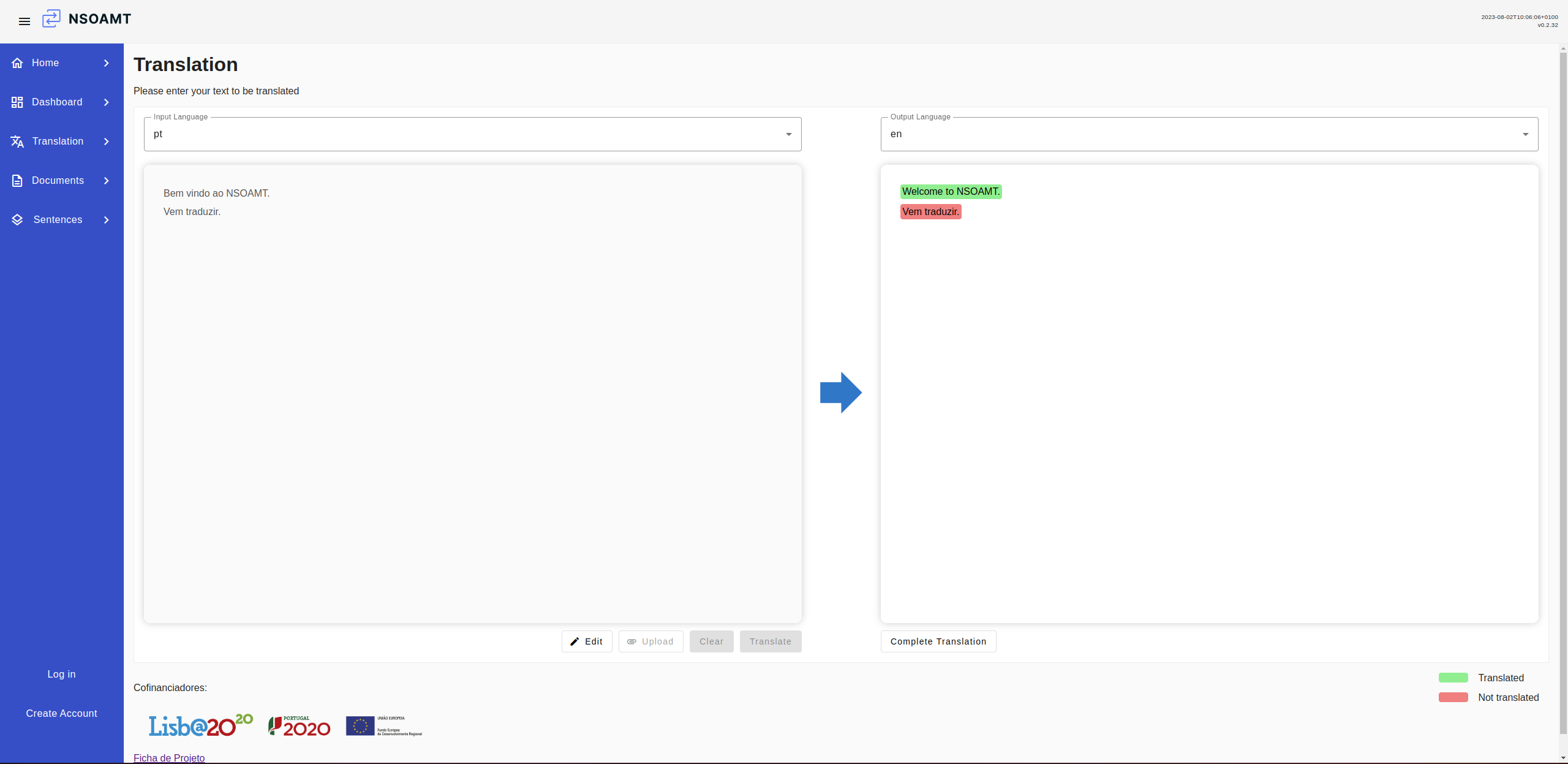}
	\caption{Translation page.}
	\label{fig:translationpage}
\end{figure*}

\begin{figure*}[htbp]
	\centering
	\includegraphics[width=0.7\textwidth]{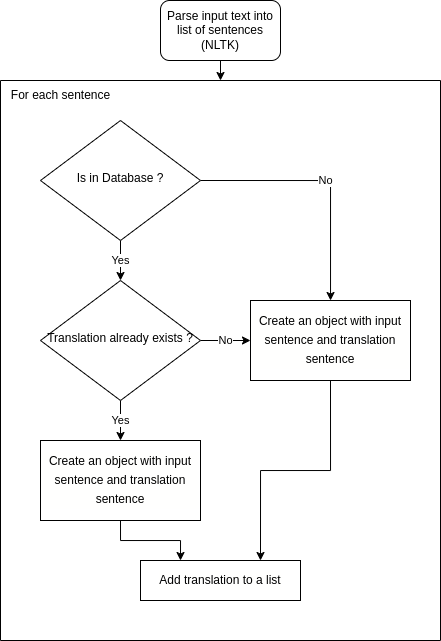}
	\caption{Translation workflow.}
	\label{fig:translationflow}
\end{figure*}

A couple of functionalities were made available in this web site. The main one is translation (See Figure \ref{fig:translationpage}). Here the user can write is own text or upload a text/plain file and translate it. The translation is made by parsing the input text into sentences and for each sentence search if already exists in the database and if already have a translation (See Figure \ref{fig:translationflow}). A list of highlighted sentences will be presented where green means translated and red means the translation was not found. 

Since the input text needs to be separated into sentences, the concept of paragraph ends up being lost. The process of reconstructing the original translated text would be complex, would require more computation and the correct reconstruction of the text is not guaranteed.
One of the main obstacles encountered is the correct processing of uploaded texts due to the wide variety of file formats. To ensure better results for the user, the file formats allowed for direct upload was restricted to text/plain, due to its simplicity and the tests made show better results on sentence division without much pre- and post-processing.

\begin{figure*}[htbp]
	\centering
	\includegraphics[width=0.7\textwidth]{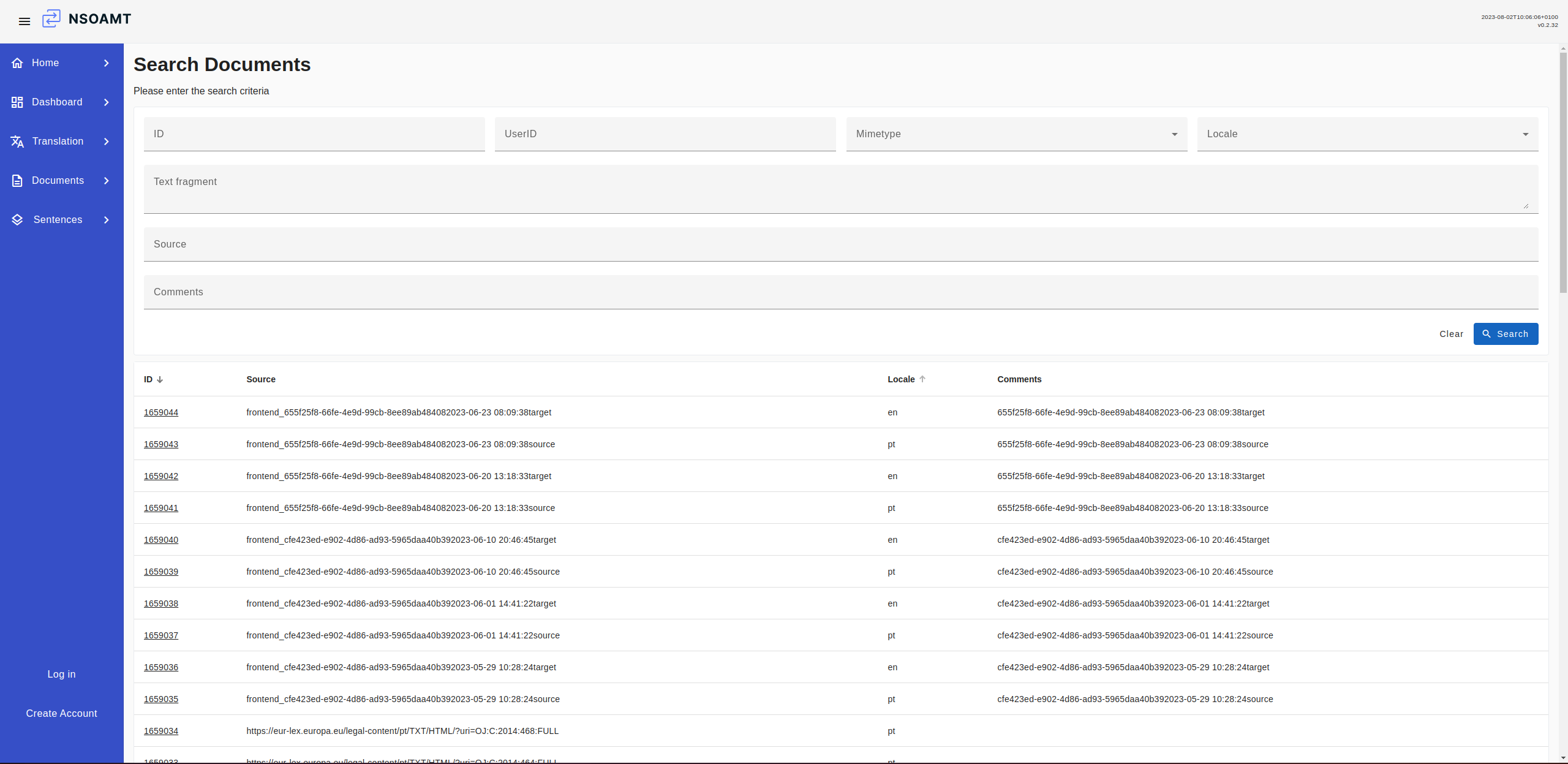}
	\caption{Search documents page.}
	\label{fig:docspage}
\end{figure*}

\begin{figure*}[htbp]
	\centering
	\includegraphics[width=0.7\textwidth]{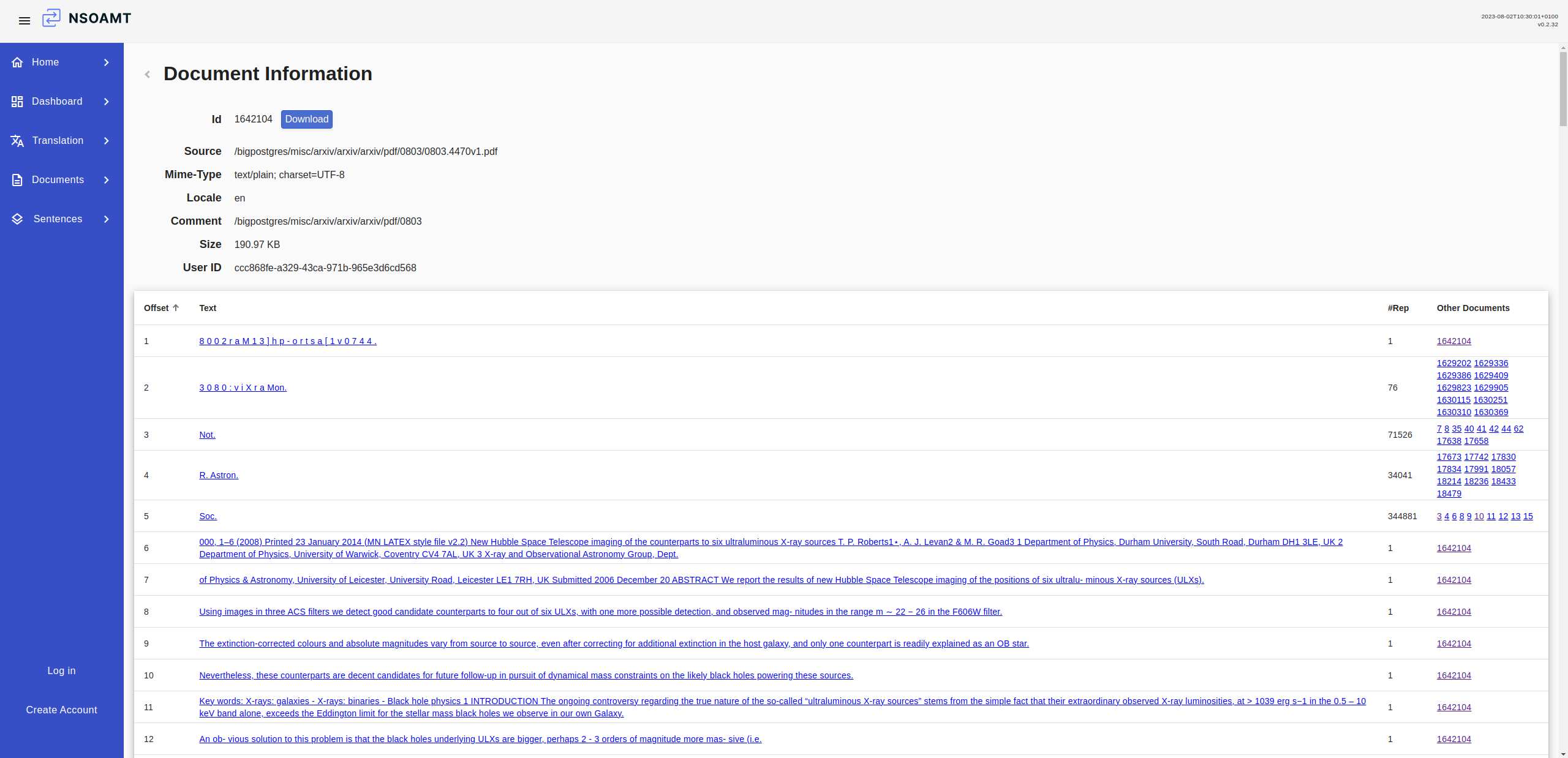}
	\caption{Document information page.}
	\label{fig:document_info}
\end{figure*}

Other functionality available on the website is to search the documents loaded into the database (See Figure \ref{fig:docspage}). From the search page it's possible to open a detailed page of a specific document (See Figure \ref{fig:document_info}). Here the user has access to more information like the MIME-type, size and if available, download the document for a complete analysis of the document. In this page is also available the list of sentences taken from the parsing process. With each sentence along comes the number of times the sentence appears in other documents and a small list of those documents. With this we can analyze the parsing process as well as the repetitions of sentences (centerpiece of this project).

\begin{figure*}[htbp]
	\centering
	\includegraphics[width=0.7\textwidth]{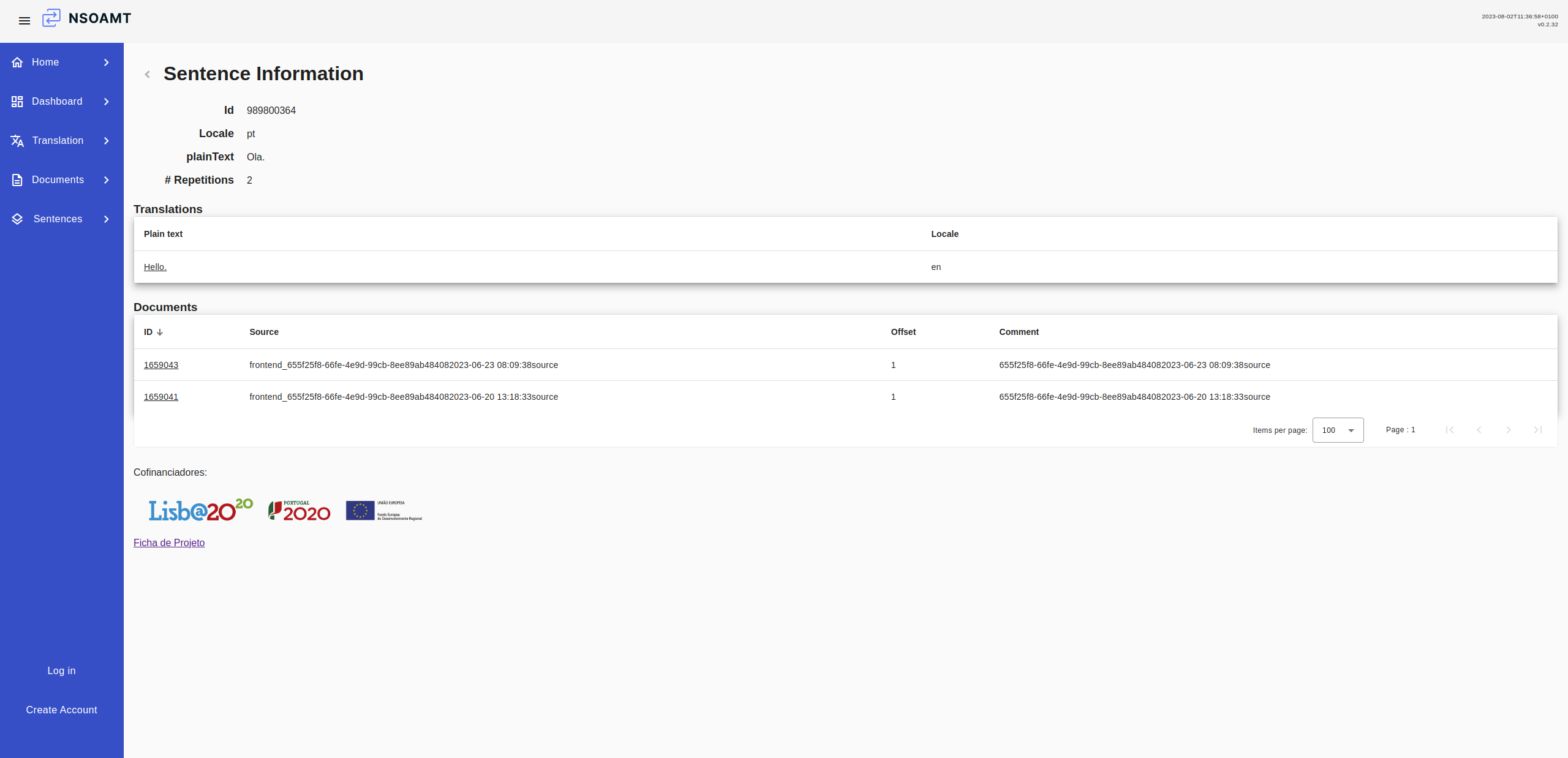}
	\caption{Sentence information page.}
	\label{fig:sentence_info}
\end{figure*}

It's also possible to search the sentences that came from the parsing process of the uploaded documents (See Figure \ref{fig:docspage}). From this page the user can go to the specific sentence page with more information about it. In the sentence information page (See Figure \ref{fig:sentence_info}) it is possible to find the number of repetitions of the sentences,  for the sentence and a list of documents where the sentence appears.

\begin{figure*}[htbp]
	\centering
	\includegraphics[width=0.7\textwidth]{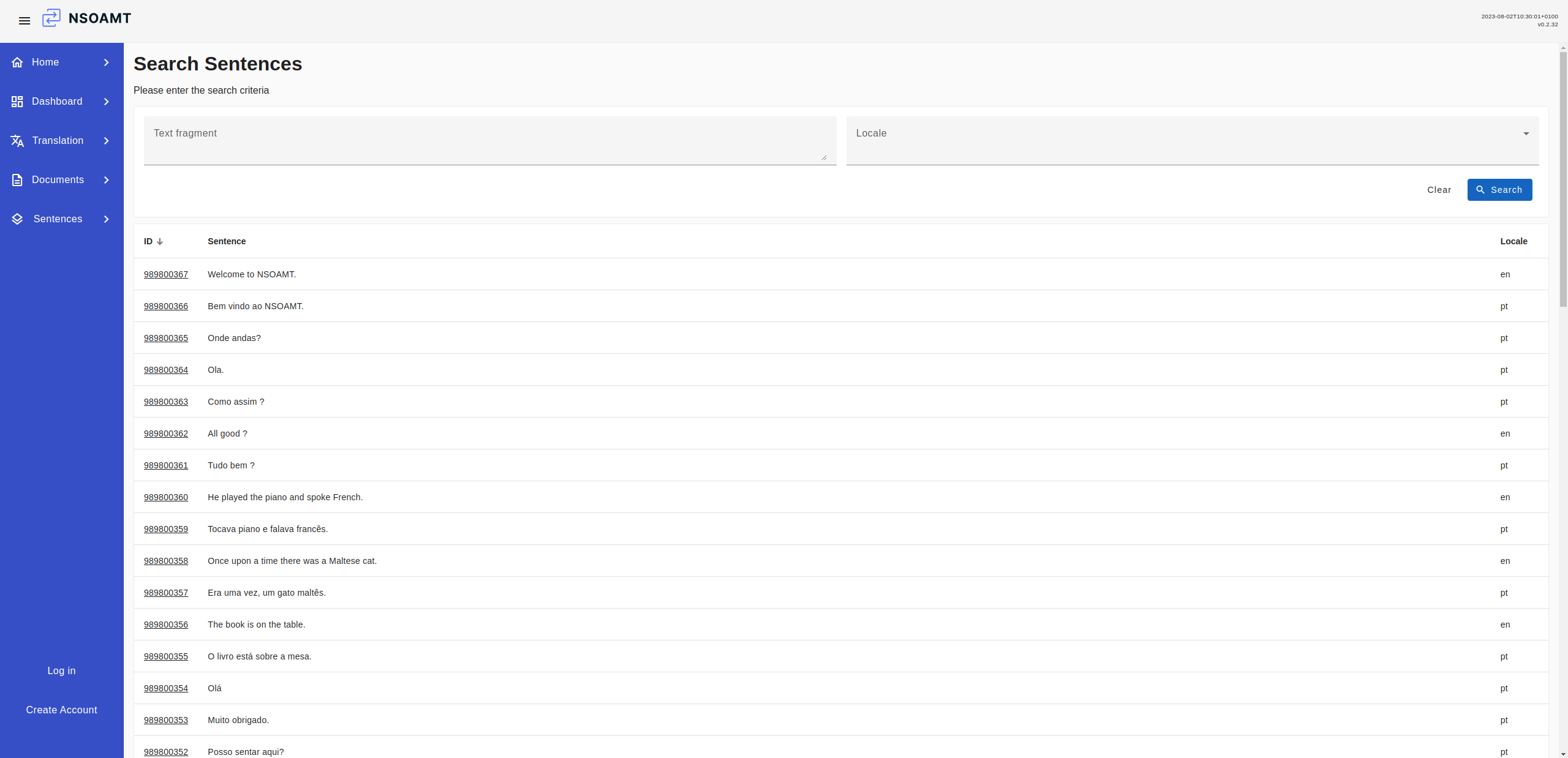}
	\caption{Search sentence page.}
	\label{fig:stcspage}
\end{figure*}

\section{Results}
\label{sec:results}

\subsection{Ingested text}
\label{sec:volume}

The total volume of text ingested is shown in Table~\ref{tab:volumefull}.

\begin{table*}[htbp]
	\centering
	\begin{tabular}{l||r|r}\hline
		& English (en) & Portuguese (pt) \\
	\hline \hline
	\#documents                        & 1,580,741       & 16 879            \\
	\#plain text characters (UTF-8) & 123,000,703,896 & 10,794,271,435     \\
	                                    & {\footnotesize $\approx$114.6 GigaBytes} & {\footnotesize $\approx$10.1 GigaBytes} \\
	\#sentences                        & 1,150,174,091   & 50,884,157         \\
	\#distinct sentences               & 885,946,972     & 23,862,499  \\
	\hline
    \end{tabular}
	\caption{Volume of text ingested}
	\label{tab:volumefull}
\end{table*}

Tables~\ref{tab:envolumestatic} and \ref{tab:ptvolumestatic} break down the volume metrics by source and language.

\begin{table*}[htbp]
	\centering
	\begin{tabular}{l||r|r|r|r}
\hline
 & ArXiv 0802-2112 & Wikipedia   (en) & EUR-LEX  (en) & tBooks   (en) \\ \hline \hline
\#documents & 1,511,891 & 61 & 17,190 & 51,593 \\
\#text characters (UTF-8) & 80,399,442,210 & 14,102,084,349 & 8,615,499,262 & 19,883,677,356 \\
\#sentences & 761,978,703 & 130,111,846 & 24,109,494 & 233,973,998 \\
 (same source) & & & & \\
\#distinct sentences & 557,332,655 & 114,423,765 & 8,112,606 & 206,490,528 \\
\#distinct sentences \% & 73.14\% & 85.29\% & 33.65\% & 88.25\% \\
 (same source) & & & & \\
\#d.sentences with repetitions & 18,914,498 & 2,426,673 & 1,712,047 & 5,471,817 \\
\#d.sentences with repetitions \% & 3.39\% & 2.12\% & 21.10\% & 2.65\% \\
 (same source) & & & & \\
\#unique d.sentences & 538,418,157 & 111,997,092 & 6 400,559 & 201,018,711 \\
\#unique d.sentences \% & 96.61\% & 97.88\% & 78.90\% & 97.35\% \\
 (same source) & & & & \\ \hline
\end{tabular}
	\caption{Volume of english text ingested from static sources}
	\label{tab:envolumestatic}
\end{table*}

\begin{table*}[htbp]
	\centering
	\begin{tabular}{l||r|r}
\hline
 & Wikipedia (pt) & EUR-LEX  (pt) \\ \hline \hline
\#documents & 8 & 16,864 \\
\#text characters (UTF-8) & 1,932,914,020 & 8,859,532,891 \\
\#sentences & 16,594,472 & 34,280,621 \\
 (same source) & & \\
\#distinct sentences & 14,805,146 & 9,060,610 \\
\#distinct sentences \% & 89.22\% & 26.43\% \\
 (same source) & & \\
\#d.sentences with repetitions & 255,735 & 2,070,827 \\
\#d.sentences with repetitions \% & 1.73\% & 22.86\%  \\
 (same source) & & \\
\#unique d.sentences & 14,549,411 & 6,989,783 \\
\#unique d.sentences \%  & 98.27\% & 77.14\% \\
 (same source) & & \\ \hline
\end{tabular}
	\caption{Volume of Portuguese text ingested from static sources}
	\label{tab:ptvolumestatic}
\end{table*}

\subsection{Common sentences}
\label{sec:commonfull}

\begin{table*}[htbp]
\centering
\begin{tabular}{l||r|r|r|r}
\hline
Common \#distinct sentences (en) & arXiv 0802-2112 & Wikipedia (en) & EUR-LEX (en) & tBooks \\ \hline \hline
arXiv 0802-2112 & \cellcolor[HTML]{FFFF00}761,978,703 & \cellcolor[HTML]{BFBFBF} & \cellcolor[HTML]{BFBFBF} & \cellcolor[HTML]{BFBFBF} \\ \hline
Wikipedia (en) & 46,531 & \cellcolor[HTML]{FFFF00}130,111,846 & \cellcolor[HTML]{BFBFBF} & \cellcolor[HTML]{BFBFBF} \\ \hline
EUR-LEX (en) & 5,448 & 28,130 & \cellcolor[HTML]{FFFF00}24,109,494 & \cellcolor[HTML]{BFBFBF} \\ \hline
tBooks & 63,747 & 145,199 & 4,665 & \cellcolor[HTML]{FFFF00}233,973,998 \\ \hline
\end{tabular}
\\[1mm]
\begin{tabular}{l|r}
\hline
Common \#distinct sentences (all sources) & 2,873 \\ \hline
\end{tabular}
	\caption{Common {\bf \#distinct sentences} between English sources}
	\label{tab:encommonstc}
\end{table*}

\begin{table*}[htbp]
\centering
\begin{tabular}{l||r|r}
\hline
Common \#distinct sentences (pt) & Wikipedia (pt) & EUR-LEX (pt) \\ \hline \hline
Wikipedia (pt) & \cellcolor[HTML]{FFFF00}16,594,472 & \cellcolor[HTML]{BFBFBF} \\ \hline
EUR-LEX (en) & 8,600 & \cellcolor[HTML]{FFFF00}34,280,621 \\ \hline
\end{tabular}
\\[1mm]
\begin{tabular}{l|r}
\hline
Common \#distinct sentences (all sources) & 8,600 \\ \hline
\end{tabular}
	\caption{Common {\bf \#distinct sentences} between Portuguese sources}
	\label{tab:ptcommonstc}
\end{table*}

Tables~\ref{tab:encommonstc} and \ref{tab:ptcommonstc} displays the number of common distinct sentences between document sources.

\subsection{Evolution of d.sentences with repetitions}
\label{sec:evoldstcrep}

A question is raised "How much volume of text would need to be ingested to have a desired \% of distinct sentences with repetitions?"

To project an answer to this question, we analyzed the evolution of metrics on the arXiv data source alone, for the following reasons:
\begin{enumerate}
    \item avoid mixing writing styles too much (between data sources).
    \item arXiv is the largest volume data source, organized by monthly folder groups.
\end{enumerate}

\begin{figure*}[htbp]
	\centering
	\includegraphics[width=0.7\textwidth]{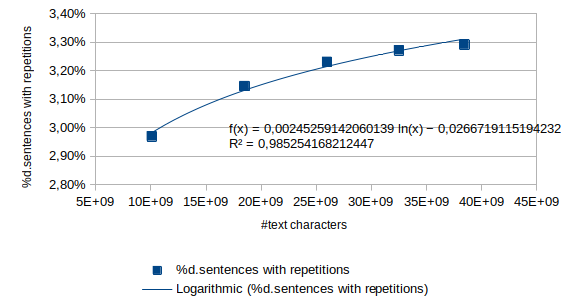}
	\caption{Trend line for \%d.sentences with rep. vs \#text characters.}
	\label{fig:arxivtrenddstcwrep}
\end{figure*}

Using the arXiv data source segmented by years 2016 to 2020, Table~\ref{tab:arxivdatadstcwrep}, a trend line was elaborated in Figure~\ref{fig:arxivtrenddstcwrep}\footnote{The trend line was calculated using LibreOffice's trend line feature, for the data points shown here.}

\begin{table*}[htbp]
	\centering
\begin{tabular}{l||r|r|r|r}
\hline
arXiv                    & \#text characters & \#d.sentences & \#d.sentences    & \%d.sentences \\
by year                  &                   &              & with repetitions & with repetitions \\ \hline \hline
2020                     & 10,076,799,973    & 72,042,632    & 2,139,053                      & 2.97\%                         \\ \hline
2019+2020                & 18,498,004,627    & 130,817,330   & 4,114,811                      & 3.15\%                         \\ \hline
2018+2019+2020           & 25,986,041,152    & 182,644,683   & 5,900,194                      & 3.23\%                         \\ \hline
2017+2018+2019+2020      & 32,503,697,718    & 227,734,087   & 7,448,879                      & 3.27\%                         \\ \hline
2016+2017+2018+2019+2020 & 38,441,439,656    & 268,911,716   & 8,849,748                      & 3.29\%                         \\ \hline
\end{tabular}
	\caption{arXiv text metrics for years 2016 to 2020.}
	\label{tab:arxivdatadstcwrep}
\end{table*}

The $R^{2}=0.985$ gives some confidence on the logarithmic trend, at least for interpolation. (Although not shown here for brevity, the logarithmic trend line was the best fit compared to other curves (linear, exponential, polynomial, etc.).\footnote{Also not shown here for brevity, but subdividing the same yearly arXiv data into months - and thus having 12 times more data points - was also visually consistent with the shown the logarithmic trend.})

Using the trend line for extrapolation, (assuming that the trend line would not change the type of curve), Table~\ref{tab:arxivprojdstcwrep} shows the projected \#text characters that would be required to achieve the desired \%d.sentences with repetitions.

\begin{table*}[htbp]
	\centering
\begin{tabular}{l||r|r|r|r|r}
\hline
\%d.sentence with repetitions & {\bf 5.00\%}   & 25.00\%  & 50.00\%  & 75.00\%   & 100.00\%  \\ \hline \hline
\#text characters             & {\bf 3.77E+13} & 9.81E+48 & 1.82E+93 & 3.39E+137 & 6.29E+181 \\ \hline
\end{tabular}
	\caption{arXiv projections for \%d.sentences with repetitions vs \#text characters.}
	\label{tab:arxivprojdstcwrep}
\end{table*}

Even for the 5\% objective (the nearest to the current 3.39\% shown in Table~\ref{tab:envolumestatic}), it does not seem practical to gather 3.77E+13 characters ($\approx$ 37 TeraBytes) of text in a short notice, to verify this projection.

But the projection is still verifiable on the total number of arXiv characters ingested. From Table~\ref{tab:envolumestatic} for 80,399,442,210 text characters:
\begin{description}
    \item[Predicted:] 3.49\% using the trend line.
    \item[Observed:] 3.39\% distinct sentences with repetitions
\end{description}

Projections for higher \%d.sentence with repetitions  (25\%, 50\%, 75\% and 100\%) are also shown in table \ref{tab:arxivprojdstcwrep} for curiosity, and should not be taken very seriously, as it is well known that for a logarithmic curve, small variations on the curve coefficient will cause large changes for distant points. These projections also assume that curve does not change shape.

\subsection{Evolution of d.v.sentences with repetitions}
\label{sec:evoldvstcrep}

This motivated the use of {\tt LanguageTool} \ref{sec:languagetool} to reduce the {\em noise level} in sentences and analyze how it would affect projections. ("v" in "d.v.sentences" stands for "valid").

On Table~\ref{tab:enarxivfullv} we compare the full arXiv metrics (already displayed on Table~\ref{tab:envolumestatic}) to the same metrics using only validated sentences.

\begin{table*}[htbp]
\centering
\begin{tabular}{l||rr}
\hline
arXiv 802-2112                        & \multicolumn{2}{r|}{sentence vs valid sentence} \\ \hline \hline
\#documents                           & \multicolumn{2}{r}{1,511,891}                                  \\ \hline
\#text character                      & \multicolumn{2}{r}{80,399,442,210}                             \\ \hline
\#sentences                           & \multicolumn{1}{r|}{761,978,703}          & 387,476,668         \\ \hline
\#d.sentences                         & \multicolumn{1}{r|}{557,332,655}          & 262,628,255         \\ \hline
\%d.sentences                         & \multicolumn{1}{r|}{73.14\%}             & 67.78\%             \\ \hline
\#d.sentences with repetitions        & \multicolumn{1}{r|}{18,914,498}          & 13,613,838         \\ \hline
\#unique d.sentences                  & \multicolumn{1}{r|}{538,418,157}         & 249,014,417             \\ \hline
\%unique d.sentences                  & \multicolumn{1}{r|}{96.61\%}             & 94.82\%             \\ \hline
\end{tabular}
    \caption{Full arXiv metrics comparing sentences with validated sentences.}
	\label{tab:enarxivfullv}
\end{table*}

\begin{figure*}[htbp]
	\centering
	\includegraphics[width=0.7\textwidth]{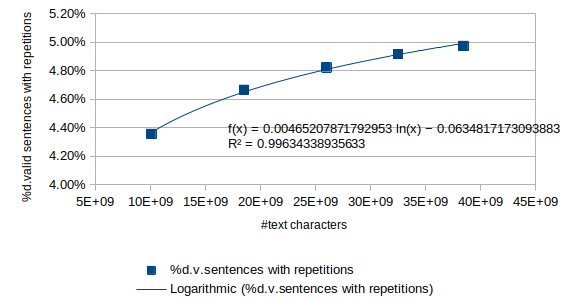}
	\caption{Trend line for \%d.sentences with rep. vs \#text characters.}
	\label{fig:arxivtrenddvstcwrep}
\end{figure*}

The data for years 2016 to 2020 is shown in Table~\ref{tab:arxivdatadvstcwrep}, and the trend line plotted on Figure~\ref{fig:arxivtrenddvstcwrep}.

\begin{table*}[htbp]
	\centering
	\tabcolsep=0.11cm
	\begin{tabular}{l||r|r|r|r|r}
\hline
arXiv                    & \#text              & \#d.sentences & \#d.v.sentences & \%valid & \%d.v.sentences \\
by year                  &  characters         &               &                 &                  & with repetitions \\ \hline \hline
2020                     & 10,076,799,973 & 72,042,632         & 36,257,428                 & 50.33\%          & 4.36\%                                    \\ \hline
2019+2020                & 18,498,004,627 & 130,817,330        & 64,926,392                 & 49.63\%          & 4.66\%                                    \\ \hline
2018+2019+2020           & 25,986,041,152 & 182,644,683        & 89,785,881                 & 49.16\%          & 4.82\%                                    \\ \hline
2017+2018+2019+2020      & 32,503,697,718 & 227,734,087        & 110,992,373                & 48.74\%          & 4.91\%                                    \\ \hline
2016+2017+2018+2019+2020 & 38,441,439,656 & 268,911,716        & 130,019,555                & 48.35\%          & 4.97\%                                    \\ \hline
\end{tabular}
	\caption{arXiv text metrics for years 2016 to 2020 for valid sentences.}
	\label{tab:arxivdatadvstcwrep}
\end{table*}

The column \#d.v.sentences is the number of d.sentences that pass the validation too check. The column \%valid is $=\frac{\#d.v.sentences}{\#d.sentences}\cdot{}100.00$. The validation tool seems to reject approx. 50\% of sentences. It can also be noticed that the \%d.v.sentences is higher than the \%d.sentences (on the distinct but-non-validated sentence data).

\begin{table*}[htbp]
	\centering
\begin{tabular}{l||r|r|r|r|r}
\hline
\%d.v.sentence with repetitions & {\bf 5.00\%}   & 25.00\%  & 50.00\%  & 75.00\%   & 100.00\%  \\ \hline \hline
\#text characters             & {\bf 3.93E+10} & 1.84E+29 & 4.02E+52 & 8.76E+75 & 1.91E+99 \\ \hline
\end{tabular}
	\caption{arXiv projections for \%d.v.sentences with repetitions vs \#text characters.}
	\label{tab:arxivprojdvstcwrep}
\end{table*}

The trend line predicts the need for 3.93E+10 ($\approx$ 28.2 GigaBytes of text) to achieve 5\% distinct valid sentences with repetitions. (Table~\ref{tab:arxivprojdvstcwrep} shows text volume extrapolations based on the same trend line for higher percentages).

From Table~\ref{tab:enarxivfullv} for 80,399,442,210 text characters ($\approx$ 8.03E+10 text characters) ($\approx$ 74.9 GigaBytes of text), we observe 5.18\% distinct valid sentences with repetitions:
\begin{description}
    \item[Predicted:] 5.33\% using the trend line.
    \item[Observed:] 5.18\% distinct valid sentences with repetitions
\end{description}

\section{Conclusions}
\label{sec:conclusions}

Some success was achieved in extrapolating the evolution of the number of distinct sentences with repetitions vs the volume of ingested text, using a trend line of logarithmic nature.

The discouraging aspect is that assuming that the trend line does not change its curve nature (from logarithmic nature to something else) at an hypothetical inflection point, it will not be practical to gather enough text volume even for modest repetition coverage (like 50\%). Not enough text volume was gathered to show evidence that this hypothetical inflection point may exist. Also, (at these large, extrapolated text volumes) it would not be feasible to crowd-source translations.

\section{Future work}
\label{sec:futurework}

The study showed interesting results for the text analyzes and translation, but one key point needs to be resolved before further work: The projections based on the current sentence string model show that it should not be possible to gather enough text documents for modest translation coverage (and the volumes needed would also be to high for effective crowd-sourcing anyway). 

Can a different sentence model provide higher rates of common text matching?

Preliminary experiments using character-level simplification techniques within a sentence (elimination of punctuation, digits, date tagging, custom sentence tokenizer, etc) have shown residual improvements that where not considered qualitatively significant to be show here.

Can a combination of techniques (such as:
\begin{itemize}
    \item Syntax trees with sub-tree matching.
    \item Take inspiration from transformers and attention models (such as other neural-network techniques \cite{vaswani2023attention}).
\end{itemize}
) be mixed with such an approach ? And will crowd-sourcing the translations of the most common text structures still be viable?


\begin{thebibliography}{10}

	\bibitem{wang2021progress}
	Haifeng Wang, Hua Wu, Zhongjun He, Liang Huang, and Kenneth~Ward Church.
	\newblock Progress in machine translation.
	\newblock {\em Engineering}, 2021.
	
	\bibitem{wiesmann2019machine}
	Eva Wiesmann.
	\newblock Machine translation in the field of law: A study of the translation
	  of italian legal texts into german.
	\newblock {\em Comparative Legilinguistics}, 37(1):117--153, 2019.
	
	\bibitem{jinfang2023exploring}
	Yao Jinfang.
	\newblock Exploring the advantages and limitations of machine translation in
	  the performance of construction industry.
	\newblock {\em International Journal of Business and Management Invention},
	  12(6):312--318, 2023.
	
	\bibitem{enwiki:1172855974}
	{Wikipedia contributors}.
	\newblock Crowdsourcing --- {Wikipedia}{,} the free encyclopedia.
	\newblock
	  \url{https://en.wikipedia.org/w/index.php?title=Crowdsourcing&oldid=1172855974},
	  2023.
	\newblock [Online; accessed 5-September-2023].
	
	\bibitem{johri2021natural}
	Prashant Johri, Sunil~K Khatri, Ahmad~T Al-Taani, Munish Sabharwal, Shakhzod
	  Suvanov, and Avneesh Kumar.
	\newblock Natural language processing: History, evolution, application, and
	  future work.
	\newblock In {\em Proceedings of 3rd International Conference on Computing
	  Informatics and Networks: ICCIN 2020}, pages 365--375. Springer, 2021.
	
	\bibitem{enwiki:1173873917}
	{Wikipedia contributors}.
	\newblock Natural language processing --- {Wikipedia}{,} the free encyclopedia,
	  2023.
	\newblock [Online; accessed 5-September-2023].
	
	\bibitem{enwiki:1173840397}
	{Wikipedia contributors}.
	\newblock Large language model --- {Wikipedia}{,} the free encyclopedia.
	\newblock
	  \url{https://en.wikipedia.org/w/index.php?title=Large_language_model&oldid=1173840397},
	  2023.
	\newblock [Online; accessed 5-September-2023].
	
	\bibitem{enwiki:1172581333}
	{Wikipedia contributors}.
	\newblock Moore's law --- {Wikipedia}{,} the free encyclopedia.
	\newblock
	  \url{https://en.wikipedia.org/w/index.php?title=Moore%27s_law&oldid=1172581333},
	  2023.
	\newblock [Online; accessed 5-September-2023].
	
	\bibitem{enwiki:1171220477}
	{Wikipedia contributors}.
	\newblock Utf-8 --- {Wikipedia}{,} the free encyclopedia.
	\newblock
	  \url{https://en.wikipedia.org/w/index.php?title=UTF-8&oldid=1171220477},
	  2023.
	\newblock [Online; accessed 21-August-2023].
	
	\bibitem{devlin2021longest}
	Thomas~Moore Devlin.
	\newblock What is the longest word in the world?
	\newblock
	  \url{https://www.babbel.com/en/magazine/the-longest-word-in-the-world}, 2021.
	
	\bibitem{rgdisc}
	Glenn Bingham.
	\newblock Are there infinitely many possible sentences in a natural language?
	\newblock
	  \url{https://www.researchgate.net/post/Are\_there\_infinitely\_many\_possible\_sentences\_in\_a\_natural\_language/532507bcd5a3f2bd398b4681/citation/download},
	  03 2014.
	
	\bibitem{levin2020longest}
	Nancy Levin.
	\newblock 10 longest known sentences in english.
	\newblock \url{https://largest.org/culture/sentences-in-english/}, 2020.
	
	\bibitem{vincent2014length}
	Sara Vincent.
	\newblock Sentence length: why 25 words is our limit.
	\newblock
	  \url{https://insidegovuk.blog.gov.uk/2014/08/04/sentence-length-why-25-words-is-our-limit/},
	  2014.
	
	\bibitem{enwiki:1170308086}
	{Wikipedia contributors}.
	\newblock General service list --- {Wikipedia}{,} the free encyclopedia.
	\newblock
	  \url{https://en.wikipedia.org/w/index.php?title=General_Service_List&oldid=1170308086},
	  2023.
	\newblock [Online; accessed 18-August-2023].
	
	\bibitem{enwiki:1156325155}
	{Wikipedia contributors}.
	\newblock New general service list --- {Wikipedia}{,} the free encyclopedia.
	\newblock
	  \url{https://en.wikipedia.org/w/index.php?title=New_General_Service_List&oldid=1156325155},
	  2023.
	\newblock [Online; accessed 18-August-2023].
	
	\bibitem{nawl}
	The academic word list.
	\newblock \url{https://www.wgtn.ac.nz/lals/resources/academicwordlist}.
	
	\bibitem{tsl}
	The toeic service list.
	\newblock \url{https://www.newgeneralservicelist.com/toeic-service-list}.
	
	\bibitem{bsl}
	The business service list.
	\newblock \url{https://www.newgeneralservicelist.com/business-service-list}.
	
	\bibitem{enwiki:1171105697}
	{Wikipedia contributors}.
	\newblock Md5 --- {Wikipedia}{,} the free encyclopedia.
	\newblock
	  \url{https://en.wikipedia.org/w/index.php?title=MD5&oldid=1171105697}, 2023.
	\newblock [Online; accessed 21-August-2023].
	
	\bibitem{enwiki:1171091464}
	{Wikipedia contributors}.
	\newblock Hash collision --- {Wikipedia}{,} the free encyclopedia.
	\newblock
	  \url{https://en.wikipedia.org/w/index.php?title=Hash_collision&oldid=1171091464},
	  2023.
	\newblock [Online; accessed 21-August-2023].
	
	\bibitem{wikiextractor}
	{Giuseppe Attardi (attardi@di.unipi.it), University of Pisa; Antonio Fuschetto
	  (fuschett@di.unipi.it), University of Pisa}.
	\newblock Wikiextractor.
	\newblock \url{https://github.com/apertium/WikiExtractor/}.
	
	\bibitem{enwiki:1171458148}
	{Wikipedia contributors}.
	\newblock Html --- {Wikipedia}{,} the free encyclopedia.
	\newblock
	  \url{https://en.wikipedia.org/w/index.php?title=HTML&oldid=1171458148}, 2023.
	\newblock [Online; accessed 21-August-2023].
	
	\bibitem{beautifulsoup}
	{Leonard Richardson}.
	\newblock beautifulsoup4 python library.
	\newblock \url{https://pypi.org/project/beautifulsoup4/}.
	
	\bibitem{enwiki:1171241740}
	{Wikipedia contributors}.
	\newblock Pdf --- {Wikipedia}{,} the free encyclopedia.
	\newblock
	  \url{https://en.wikipedia.org/w/index.php?title=PDF&oldid=1171241740}, 2023.
	\newblock [Online; accessed 21-August-2023].
	
	\bibitem{pdfminer}
	{pdfminor.six community}.
	\newblock pdfminor.six python library.
	\newblock \url{https://pypi.org/project/pdfminer.six/}.
	
	\bibitem{yu2020extracting}
	Changfeng Yu, Cheng Zhang, and Jie Wang.
	\newblock Extracting body text from academic pdf documents for text mining.
	\newblock \url{https://arxiv.org/abs/2010.12647}, 2020.
	
	\bibitem{rama2012}
	{Ramakrishnan, C., Patnia, A., Hovy, E. et al.}
	\newblock Layout-aware text extraction from full-text pdf of scientific
	  articles.
	\newblock \url{https://doi.org/10.1186/1751-0473-7-7}, 2012.
	
	\bibitem{nltk}
	Natural language toolkit.
	\newblock \url{https://www.nltk.org/}.
	
	\bibitem{ltool}
	Language tool.
	\newblock \url{https://languagetool.org/}.
	
	\bibitem{vaswani2023attention}
	Ashish Vaswani, Noam Shazeer, Niki Parmar, Jakob Uszkoreit, Llion Jones,
	  Aidan~N. Gomez, Lukasz Kaiser, and Illia Polosukhin.
	\newblock Attention is all you need.
	\newblock \url{http://arxiv.org/abs/1706.03762}, 2017.
	
	\end{thebibliography}
\end{document}